\documentclass[letterpaper, journal]{IEEEtran}

\usepackage[noadjust,space]{cite}
\usepackage{svg}
\usepackage{amsmath,amssymb,amsfonts}
\usepackage{algorithmic}
\usepackage{graphicx}
\usepackage{textcomp}
\usepackage{xcolor}
\usepackage{tabularx}
\usepackage{dsfont}
\usepackage[]{subcaption}
\usepackage{colortbl} 
\usepackage{hhline} 
\usepackage[colorlinks=true,allcolors=.,urlcolor=blue,]{hyperref}
\usepackage{multicol}
\usepackage{flushend}
\usepackage{multirow}
\usepackage{epsfig}

\def\tablename{Table}
\DeclareCaptionLabelFormat{mytablestyle}{\tablename~\thetable}
\renewcommand{\thetable}{\arabic{table}}

\DeclareMathOperator*{\argmax}{\arg\!\max}

\graphicspath{ {./images/} }

\makeatletter
\def\convertto#1#2{\strip@pt\dimexpr #2*65536/\number\dimexpr 1#1}
\makeatother

\title{\LARGE \bf
	Multimodal Visual-haptic pose estimation in the presence of transient occlusion
}

\author{Michael Zechmair$^{1}$, Alban Bornet$^{2}$  and Yannick Morel$^{1}$
	\thanks{$^{1}$Michael Zechmair and Yannick Morel are with Faculty of Psychology and Neuroscience, Maastricht University, %
		Maastricht, The Netherlands, %
		{\tt\small \{m.zechmair,y.morel\}@unimaas.nl}}%
	\thanks{$^{2}$Alban Bornet is with with Department of Radiology and Medical Informatics, University of Geneva, Geneva, Switzerland, {\tt \small alban.bornet@unige.ch} and was with Laboratory of Psychophysics, Brain Mind Institute, EPFL, Lausanne, Switzerland}%
	\thanks{This research has received funding from the European Union's Horizon 2020 Framework Programme for Research and Innovation under the Specific Grant Agreement No. 785907 (Human Brain Project SGA3).}%
}

\begin{document}
	\newcolumntype{a}{>{\columncolor{lightgray}}c}
	
	\maketitle
	\begin{abstract}
Human-robot collaboration requires the establishment of methods to guarantee the safety of participating operators. A necessary part of this process is ensuring reliable human pose estimation. Established vision-based modalities encounter problems  when under conditions of occlusion.
This article describes the combination of two perception modalities for pose estimation in environments containing such transient occlusion. We first introduce a vision-based pose estimation method, based on a deep Predictive Coding (PC) model featuring robustness to partial occlusion. Next, capacitive sensing hardware capable of detecting various objects is introduced. The sensor is compact enough to be mounted on the exterior of any given robotic system. The technology is particularly well-suited to detection of capacitive material, such as living tissue. Pose estimation from the two individual sensing modalities is combined using a modified Luenberger observer model. We demonstrate that the results offer better performance than either sensor alone. The efficacy of the system is demonstrated on an environment containing a robot arm and a human, showing the ability to estimate the pose of a human forearm under varying levels of occlusion.
	\end{abstract}

	\section{Introduction}

	In recent decades, robotic applications have become prevalent in industrial processing and manufacturing lines. Robots have been used to automate tasks and reduce human workloads. In particular, robots are well suited to executing repeatable, well-defined processes. Recent research has focused on expanding the capabilities of robotic applications with the aim of creating more autonomous systems that can adapt to environmental factors that alter task parameters (\cite{schuh2017}). As these robots could be deployed in complex production lines, more processes could be automated. 
To ensure reliable interactions with their surroundings, robotic applications require robust methods of environmental characterization. This includes detecting and localizing nearby objects of interest, particularly those necessary to perform a given task.
One approach is to develop methods of human-robot collaboration, as the capabilites of humans and robots synergize well. Humans are capable of adapting to changes in established processes, and can adjust a robot's behavior accordingly. However, having humans and robots collaborate on a task requires the establishment of safety guarantees to prevent injuring humans. 
As such, precise, real-time localization is fundamental for robots to understand human intentions, adapt to dynamic behaviors, avoid collisions, and effectively collaborate with their human counterparts.

Current approaches to object and human localization heavily focus on vision-based methods. This relies on segmenting a vision-based input, such as an RGB camera, highlighting how various objects are distributed in an image.
Most modern computer vision methods use convolutional neural networks (CNNs) to solve vision segmentation tasks (\cite{chen2017deeplab, chen2018encoder, girshick2014rich, he2017mask, long2015fully, ronneberger2015u}). Due to their versatility, CNNs are also the most performant models for many other vision tasks, including image recognition (\cite{he2016deep, krizhevsky2012imagenet, tan2019efficientnet}), image generation (\cite{goodfellow2014generative, karras2019style, radford2015unsupervised, zhu2017unpaired}), and scene rendering (\cite{eslami2018neural, sakaridis2018semantic, xiao2018unified}), among others. Because of this flexibility, CNNs have proven useful for a wide range of image-based practical applications such as medical imaging (\cite{shen2015multi, shin2016deep}), autonomous driving (\cite{xu2017end}), or manufacturing (\cite{han2019novel, sateesh2016deep, wen2020transfer}). However, it has been observed that CNNs face problems when encountering visual occlusion (\cite{economist2017uber, fawzi2016measuring, kortylewski2020combining, mcallister2017concrete}). In the case of visual segmentation, image contours do not support the boundaries of an occluded object, whereas the occluder’s contours may be mistaken for the boundaries of a considered object. As discussed in \cite{zhu2022challenges}, this problem is particularly significant in collaborative robotic (“cobotic”) use cases, in which humans and robots work together in close proximity. Such interactions lead to strong and frequent occlusions, heavily impairing inferences made by the network.

This challenge is very relevant to everyday human vision: despite encountering elements that appear very rarely in isolation, human vision is notably robust to occlusion (\cite{rajaei2019beyond, zhu2019robustness, lamme2002masking, tang2018recurrent, wyatte2012limits}). Hence, using insights from human vision may grant the keys to increasing CNNs’ robustness to occlusions. More in general, taking inspiration from the human visual cortex may help improve the flexibility and generalization capabilities of neural networks. Importantly, although CNNs reach human-level performance in many complex vision tasks and are the best models of image-evoked population response in the primate visual cortex (\cite{kriegeskorte2015deep, schrimpf2018brain, yamins2014performance}), fundamental differences exist between humans and CNNs. Indeed, human-like performance of CNNs does not necessarily imply human-like computations. For example, the role of feedback connections in the human visual cortex remains a matter of debate (\cite{olshausen2006other, van2020going}). Understanding these discrepancies is crucial to move towards more human-like models. To this end, there exists a large corpus of results from vision psychophysics paradigms that CNNs cannot explain.
For example, visual crowding experiments have shown that the human visual cortex integrates information across very large portions of the visual field (\cite{malania2007grouping, saarela2009global, vickery2009supercrowding}). These experiments suggest that, in human vision, high-level context about the global configuration of the visual input strongly affects local and low-level information processing (\cite{manassi2012grouping, manassi2013crowding, manassi2016crowding}). In contrast, CNNs cannot reproduce these results because they are based on feedforward and local operations only (\cite{bornet2021global, doerig2020crowding}). Identifying what is missing from CNNs to account for global aspects of crowding is a good way to understand differences to human-like computations. Recent studies showed that the only models of human vision explaining the global aspects of visual crowding include explicit recurrent grouping and segmentation processes (\cite{bornet2021shrinking, doerig2019beyond}). For example, adding dynamic routing to CNNs (capsule networks) (\cite{sabour2017dynamic}) or illusory contour mechanics (\cite{francis2017neural}), both of which instantiate grouping and segmentation, matched human behaviour in visual crowding paradigms (\cite{bornet2021shrinking, choung2021dissecting, doerig2020capsule}). This line of evidence suggests that one computational function of recurrent processing in the human visual cortex is to efficiently select which features integrate (grouping) and which do not (segmentation). These processes help the brain cope with complex input data (occlusions, reflections, noise, etc.) and refine low-level information based on high-level context. In computer vision for example, adding feedback processing to CNNs is helpful to perform inferences from partial information (\cite{rajaei2019beyond, tang2018recurrent, linsley2020recurrent, otto2006flight}).

We built upon such psycho-physical insights to create a visual segmentation model providing a degree of robustness to occlusion. More specifically, we present a method able to learn to produce segmentation masks based on Predictive Coding, and infer likeliest pose by comparing the segmentation mask produced by the PC model
to a range of candidate masks, reflecting different candidate poses: ProcNet.

While vision-based localization has been extensively employed for its non-intrusiveness and versatility, it faces significant challenges in environments where occlusion is prevalent. Cluttered spaces or close interactions can obstruct visual cues, leading to incomplete or inaccurate localization. Human-robot collaboration in particular requires close interactions, leading to occlusion. The limitations of vision-based methods necessitate the exploration of multimodal localization approaches which address occlusion-dependent inaccuracies and enhance overall localization accuracy.
Embracing multimodal approaches that fuse data from multiple sensors enables robots to compensate for the limitations of individual sensors and achieve more comprehensive and robust localization, particularly in challenging environments.
Among various sensing alternatives, capacitive sensing has emerged as an ideal additional modality for enhancing human localization in robotic applications. By leveraging changes in electrical fields, capacitive sensors can accurately detect and track nearby objects, including humans. However, the sensor's limited range necessitates a complementary, broad-range sensing modality. Therefore, combining capacitive and vision-based localization allows us to cover the entire volume of a robot's working environment. By placing capacitive sensors along the robot's hull, we can localization nearby objects at close range, while vision-based sensing localizes objects in a broad area in the workspace.

Capacitive sensors have been used to support detection of a broad range of  materials. In \cite{reverter2007liquid}, the authors present a simple sensor design to measure the level of liquid present in a container, requiring only two metal rods. Alternatively, the water content in crude-oil can be accurately determined by measuring sensor phase changes (\cite{aslam2014high}). Water flow rates can be measured by similar techniques, as shown in \cite{chiang2006semicylindrical}. A novel sensor electrode design intended to improve sensitivity to humidity is presented by the authors of \cite{rivadeneyra2014novel}. Very precise distance measurements of metallic objects have been performed by combining capacitive with inductive-based sensors (\cite{kan2018dual}). The technology can also be used to for wearables, supporting the design of flexible sensors allowing to quantify human activity. Variables measured include respiratory and heart rate, as well as muscle deformation (\cite{cheng2010active,kundu2013wearable}). Capacitive sensors have also been employed to measure a number of object properties. In \cite{fulmek2002capacitive}, the authors describe the implementation of a sensor on two rotors allowing the determination of their relative rotation angle. The approach in \cite{roberts2013soft} uses three liquid electrodes, encased in elastic material, to measure both shear and pressure deformation. As capacitive sensors mainly require two conductive plates to function, they can even be directly integrated into production processes by commercially available, multi-material 3D-printers, as discussed in \cite{shemelya20133d}.

For the purpose of human-robot interaction and accident prevention, a number of haptic-sensing based approaches have been investigated. In \cite{wrro166006}, various methods of active touch sensing used by mammals are discussed, along with their applications to robotics. In \cite{lamy2009robotic}, a pressure-sensitive skin that can detect human touch when mounted on a robot's hull is proposed. While this approach is based on resistance measurements, the authors in \cite{vsekoranja2014human} demonstrate an analogous sensor with the same touch-sensing capabilities, based on capacitive sensing. They also argue that by mounting the sensors on all segments of a robot arm, contact between robot and human can be detected and the robot's movements can be immediately stopped, preventing injury. This concept is expanded on in \cite{pritchard2008flexible}, where a pressure-sensitive skin is created that employs an array of small, 1$\rm{mm}^2$ electrodes on flexible circuit boards, allowing precise measurements of applied forces. 
Expanding on the idea of safety in cobotics, the authors in \cite{goger2013tactile} demonstrate a sensor that is capable of both contact-sensing as well as close-range human-presence detection. The developed sensor is mounted on an anthropomorphic robot hand, allowing distance-aware interactions between robot and human. In \cite{xia2018tri}, a capacitive sensor array that is directly integrated into workers' clothing is proposed. The sensors are capable of detecting nearby robots and can stop those robot's movements remotely. Conversely, a multi-electrode based capacitive sensor is presented in both \cite{hoffmann2016environment} and \cite{schlegl2013virtual}, intended to be mounted on robotic segments. The latter also demonstrates a system which, by employing their capacitive sensor, is capable of detecting and stopping a moving robot arm in the presence of a human. Building on these capabilities, the author in \cite{erickson2018tracking} employs a single capacitive sensor to pull a hospital gown over a human's arm, maintaining a safe distance in the process.

To combine the short-range capabilities of capacitive sensors with the long-range abilities of vision-based systems, we use a modified Luenberger observer model. Observers are generally used for determining an estimate of the internal state of a system (\cite{khalil2015nonlinear}). Here, the relevant internal state is the pose of the tracked object, which can be inferred from the sensing modality measures. A similar approach to combining sensor measurements was demonstrated by the authors in \cite{gomez2020sensor}, where an observer model was used to estimate the pose of a quadcopter based on visual and interial sensors. Similarly, in \cite{rigatos2010extended}, an extended Kalman filter combines sensor information received by odometric and sonar sensors. Observers have been employed in a wide range of use-cases, include sensor fusion for electrochemical applications (\cite{ross2023sensor}), the charge state of lithium-ion batteries (\cite{hu2010estimation}), and to determine the pose of a robotic gripper (\cite{celani2006luenberger}).

The contribution of this article is a multimodal sensing modality for object pose estimation, containing novel vision- and haptic-based sensing modalities aimed at minimizing the negative effects of occlusion. We present methods for object localization based on both haptic sensing and vision. For haptic sensing, we present a compact capacitive-based hardware system that can be mounted on a robot arm, and a method to evaluate sensor output to estimate a considered object's pose. As stated above, ProcNet (our approach to vision-based localization) is based on a predictive-coding segmentation method, combined with a custom object pose estimation model that compares 3D-object shapes to a segmentation mask. Results of the two pose estimation methods are evaluated and combined via a modifed Luenberger-type observer model, enabling occlusion-resistant localization. Lastly, we compare performance to another established pose estimation model, NVIDIA's PoseCNN (\cite{xiang2017posecnn}).

The following article is organized as follows: Section \ref{sec:perception_modalities} describes our perception modalities, with section \ref{sec:active_electric_perception} focusing on active electric perception, and section \ref{sec:procnet} on ProcNet. Section \ref{sec:multimodal_localization} details our method of merging both approaches, and section \ref{sec:human_limb_tracking} provides an overview over results. Lastly, section \ref{sec:conclusion} concludes the article.

\section{Perception Modalities} \label{sec:perception_modalities}

	In this section, we introduce the two previously discussed localization methods based on, haptic sensing and vision. We present the two methods separately, detailing their setup, evaluating their performance, and determining their capabilities and limitations. For haptic sensing, we present the capacitive-based hardware setup and corresponding pose estimation model. For vision-based sensing, we utilize our ProcNet model for frame segmentation along with our method for subsequent object localization.

	\subsection{Active Electric Field Modality} \label{sec:active_electric_perception}
	In the following, we present the manner in which we are able to estimate the pose of electrically sensitive objects in proximity to a haptic sensor.
	
	\subsubsection{Perception Strategy}
\begin{figure}
	\centering
	\def\svgwidth{1.0\columnwidth}
\begingroup%
  \makeatletter%
  \providecommand\color[2][]{%
    \errmessage{(Inkscape) Color is used for the text in Inkscape, but the package 'color.sty' is not loaded}%
    \renewcommand\color[2][]{}%
  }%
  \providecommand\transparent[1]{%
    \errmessage{(Inkscape) Transparency is used (non-zero) for the text in Inkscape, but the package 'transparent.sty' is not loaded}%
    \renewcommand\transparent[1]{}%
  }%
  \providecommand\rotatebox[2]{#2}%
  \newcommand*\fsize{\dimexpr\f@size pt\relax}%
  \newcommand*\lineheight[1]{\fontsize{\fsize}{#1\fsize}\selectfont}%
  \ifx\svgwidth\undefined%
    \setlength{\unitlength}{252bp}%
    \ifx\svgscale\undefined%
      \relax%
    \else%
      \setlength{\unitlength}{\unitlength * \real{\svgscale}}%
    \fi%
  \else%
    \setlength{\unitlength}{\svgwidth}%
  \fi%
  \global\let\svgwidth\undefined%
  \global\let\svgscale\undefined%
  \makeatother%
  \begin{picture}(1,0.49031648)%
    \lineheight{1}%
    \setlength\tabcolsep{0pt}%
    \put(0,0){\includegraphics[width=\unitlength,page=1]{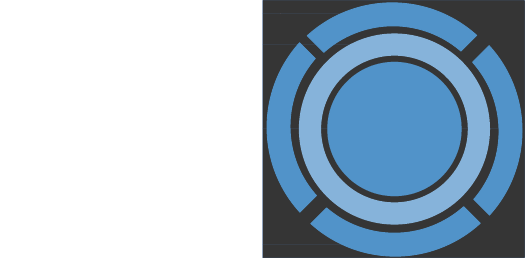}}%
    \put(0.74991998,0.23133922){\makebox(0,0)[t]{\lineheight{1.25}\smash{\begin{tabular}[t]{c}\textbf{c}\end{tabular}}}}%
    \put(0.97274972,0.23134518){\makebox(0,0)[t]{\lineheight{1.25}\smash{\begin{tabular}[t]{c}\textbf{r}\end{tabular}}}}%
    \put(0.53104739,0.23160382){\makebox(0,0)[t]{\lineheight{1.25}\smash{\begin{tabular}[t]{c}\textbf{l}\end{tabular}}}}%
    \put(0.74958141,0.44664643){\makebox(0,0)[t]{\lineheight{1.25}\smash{\begin{tabular}[t]{c}\textbf{t}\end{tabular}}}}%
    \put(0.75155734,0.00655623){\makebox(0,0)[t]{\lineheight{1.25}\smash{\begin{tabular}[t]{c}\textbf{b}\end{tabular}}}}%
    \put(0,0){\includegraphics[width=\unitlength,page=2]{sensor_board_image.pdf}}%
  \end{picture}%
\endgroup%

	\caption{Integration of electrode board with signal processing hardware (left) and diagram (right) consisting of one excitation electrode (middle, light blue ring) and five measurement electrodes (dark blue areas, \textbf{c}enter, \textbf{l}eft, \textbf{r}ight, \textbf{t}op, \textbf{b}ottom).}
	\label{fig:electrode_board}
\end{figure}

	Haptic, capacitive-based sensors function by generating an electric field and measuring how nearby objects interact with it. The sensors are composed of at least two electrodes, one to emit an electric field, and one to measure it. An example setup can be seen in Fig. \ref{fig:electrode_board}. In general, a high-frequency field is preferable, for two reasons. First, sensors that operate at higher frequencies are more resistant to external noise, as most electric field noise occurs in the lower frequency bands, at around 0-1kHz (\cite{smith1999electric}). The second benefit their ability to generate strong electric fields with simple hardware components; in general, an increase in field strength at the emitting electrode leads to a larger signal at the measurement electrode, culminating in higher SNR (Signal-to-Noise Ratio), and ultimately better perception of nearby objects. Therefore, capacitive sensors should strive to generate strong electric fields. However, the strength of an electric field is determined by the voltage applied to the emitting electrode. Generating high voltages at DC (direct current) is far more complex than at AC (alternating current). Sensor's operating at high frequencies can incorporate an inductor-capacitor circuit component (also called an LC tank) into their design to convert a low-voltage input into a high-voltage signal at their resonance frequency. This circuit is easy to design and implement, making it an ideal driver for the emitting electrode.

The sensor's measurement electrode is connected to an amplifying circuit, comprised of an opamp in gain configuration. The amplified signal is passed to the ADC (analog-digital converter) of a microcontroller (\cite{pic32mk}). We selected this specific microcontroller due to its fast onboard ADC and high operating frequencies, allowing us to process the incoming signals in real-time and compute the respective signal strength using Fourier series decomposition at operating frequency. At a sampling rate of 1kHz, our configuration can process up to 40 measurement signals simultaneously. As each sensor has a sensing range of around 15cm, we can cover a large surface area with minimal amount of hardware (for additional details, see our previous work \cite{zechmair2023active}). In our setup (see Fig. \ref{fig:robot_human_limb_environment}), we were able to detect and localize objects near a robot arm by attaching sensors on a large portion of the hull.
	
	\subsubsection{Experimental Results} \label{sec:cap_experimental_results}

\begin{figure*}
	\hspace{-0.5em}%
	\begin{subfigure}{0.5\textwidth}
		\centering
		\includegraphics[width=0.9\textwidth]{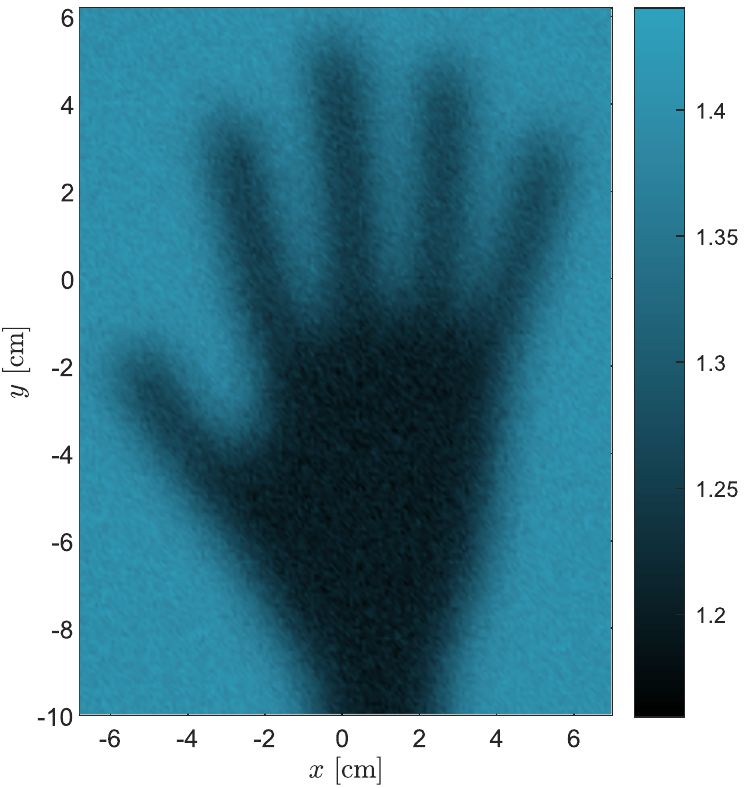}
	\end{subfigure}%
	\hfill%
	\hspace{1em}%
	\begin{subfigure}{0.5\textwidth}
		\centering
		\includegraphics[width=0.95\textwidth]{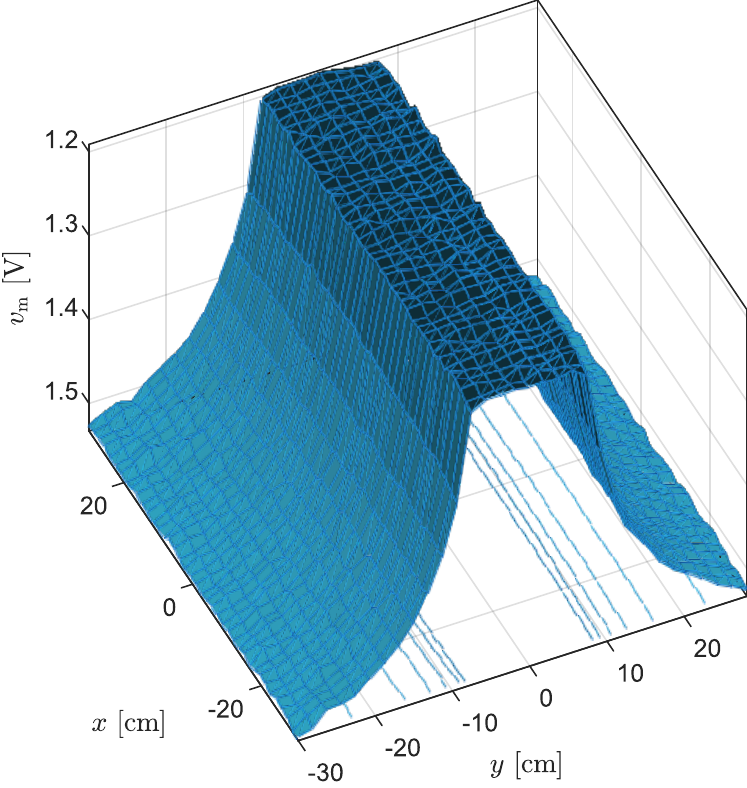}
	\end{subfigure}
	\caption{Haptic sensing measures recorded when considering a human hand (left) and forearm (right). The sensor was moved across the horizontal plane, with electrodes above the target at a constant height of 1cm.}
	\label{fig:human_limbs_scan}
\end{figure*}

To characterize the sensor's performance, a series of tests were conducted. To ensure repeatability of measures, we mounted the electrodes and electronics on the end effector of a Franka Emika Panda 6 Degree-of-Freedom (DoF) robot arm (\cite{fepanda}). The setup allows us to systematically determine electrode poses in relation to object pose when recording sensor responses.
An example of measures is shown in Fig. \ref{fig:human_limbs_scan}, where a human hand and forearm were sampled at a constant height.
Similarly, we obtained sample measures for a series of objects at various distances, ranging from 1cm to 20cm, in increments of 5mm. Sensor angles were aligned from 0° to 90° in relation to the horizontal mounting plane, in increments of 10°. All gathered samples and data is available online \href{https://github.com/Mike2208/active_electric_perception}{here}.
Gaussian regression was applied in post-processing to the measures (\cite{rasmussen2006gaussian}). Results indicate that the sensor provides a measurable reaction to the presence of nearby objects even when they are not directly in front of the electrodes. Accordingly, for the purpose of proximity detection, a single sensor could be used to monitor a large volume of space around a robot arm.

	\subsubsection{Sensor Model} \label{sec:electric_model}
	\begin{figure}
	\centering
	\def\svgwidth{1.0\columnwidth}
	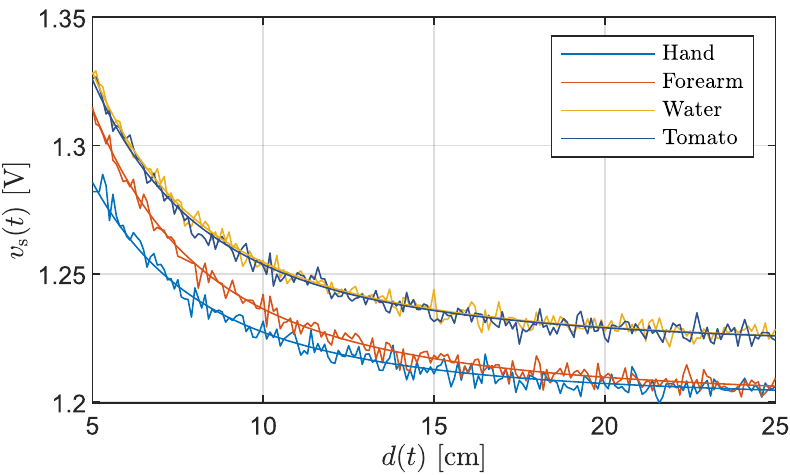
	\caption{Measures obtained with the haptic sensors above various considered objects. The robot arm was moved so that distance $d(t)$ between electrodes and object ranged from 0.5 to 20cm. The smooth lines are generated by (\ref{eq:sensor_approx}).}
	\vspace{-1.5em}
	\label{fig:dist_measurement}
\end{figure}

To evaluate sensor performance in various scenarii, we derived a model reflecting sensor behavior. As discussed in section \ref{sec:cap_experimental_results}, measures were gathered to determine the response to the presence of (and distance to) various objects. 
Resulting measures describe the relation between relative sensor-object distance and recorded voltage $v_{\text m}(t)$ (see Fig. \ref{fig:dist_measurement}).
The results show a predictable, well-behaved structure, which can be approximated using
\begin{equation}
v_{\text d}(t) = \frac{a_1}{1 + a_2 d^2(t)} + a_3, \quad t \geqslant 0, \label{eq:sensor_approx}
\end{equation}%
where $v_{\text d}(t)$ is the estimated sensor measure in V, $a_1, a_2, a_3 \in \mathds{R}^+$ are parameters describing the sensor's behavior (in V, $\text{m}^{-2}$, and V, respectively), and $d(t)$ is the distance (in $\text{m}$) between sensor center and the considered object's center of mass.
Parameters depend on both sensor configuration (surface area) as well as the object's properties (its shape, and dielectric parameter).
The SNR (Signal-to-Noise Ratio) is computed as described in \cite{PRICE199310} as
\begin{equation}
	\sigma\left(v_\text{d}(t)\right) = 10 \log_{10} \frac{\Big( \max \left(v_\text{d}(t) \right) - \min \left(v_\text{m}(t) \right) \Big)^2}{2\text{Var}[n_\text{d}(t)]}, \nonumber
\end{equation}%
where $\sigma(v_\text{d}(t)) \in \mathds{R}$ is the SNR in dB, $\text{Var}(x(t)) \in \mathds{R}^+$ is the variance of $x(t)$, and $n_\text{d}(t)$ is the noise present in $v_\text{d}(t)$ in V (with all signal noise being treated as white noise). Note that $t$ spans the entire considered measurement period. To determine $n_\text{d}(t)$, we measured the electrode output with no object present. Noise variance was assessed to be about $6.2 \text{mV}^2$.
The SNR can be used as an indicator of a sensor's detection range for various objects.
For instance, the SNR sinks under 1dB at a distance of around 15cm in the case of the human forearm. Such a detection range is useful in supporting human-robot interactions, as shown in section \ref{sec:human_limb_tracking}.
	
	\subsubsection{Haptic-based Pose Estimation}
	Our haptic sensors are capable of estimating the relative position of the closest point on the surface of a nearby object. By taking samples from multiple sensors, we can determine  various points along an objects hull. Given enough samples, it is even possible to reconstruct the shape of an object. An example is shown in Fig. \ref{fig:human_limbs_scan}, where a series of samples form the outline of a human hand.
The above figure was generated using a single sensor taking multiple measures over the course of half an hour. For our use case, only a few sensors will be in range of the  object at a given time, resulting in a limited number of samples. A complete reconstruction of an object's shape is not possible. However, a small number of samples are sufficient to estimate a known object's pose, defined as
%
%
\begin{align}
\nu_{\text{o}}(t) \triangleq %
\left[\begin{matrix}
	x_{\text{o}}(t) \\
	\eta_{\text{o}}(t)
\end{matrix}\right], \quad t \geqslant 0,
\end{align}%
where $\nu_{\text{o}}(t) \in \mathds{R}^6$ is the pose of object $o$, $x_{\text{o}}(t) \in \mathds{R}^3$ is the position of the object's center of mass in m, and $\eta_{\text{o}}(t) \in \mathds{R}^3$ the attitude using Euler coordinates expressed in radians (axis transformations executed in RPY (Roll-Pitch-Yaw) order). Each sensor delivers an estimated relative position of the nearest point of an object, as
\begin{equation}
	x_{\text{r}s}(t) = x_{\text{o}}(t) - x_{s}(t) + e_{\text{x}s}(t),
\end{equation}%
where $s \in \mathds{N}$ is the index of a single sensor, $x_{s}(t) \in \mathds{R}^3$ is the sensor's current position, and $e_{\text{x}s}(t) \in \mathds{R}^3$ is the estimation error.
All sensors are mounted on a 6-DoF (degrees-of-freedom) robot arm (\cite{fepanda}), with joint angles $\theta(t) \in \mathds{R}^6$ in radians. As the sensors are placed at fixed positions along the hull, we can use the robot's kinematics to determine $x_{s}(t)$.
However, computing an object's pose from sensor and kinematics information is not trivially possible; as the sensor samples are dependent on an object's surface, the computation requires surface analysis to determine the relation between sensor measures and relative pose. We employed a neural network, depicted in Fig. \ref{fig:cap_sensor_neural_net}, to evaluate the relationship. The network takes the robot's joint angles $\theta(t)$ and sensor measures $x_{\text{r}s}(t)$ as input, and outputs the estimated object pose $x_{\text{o}}(t)$ and $\Omega_{\text{o}}(t)$. 

\begin{figure}
	\centering
	\includegraphics[width=\columnwidth]{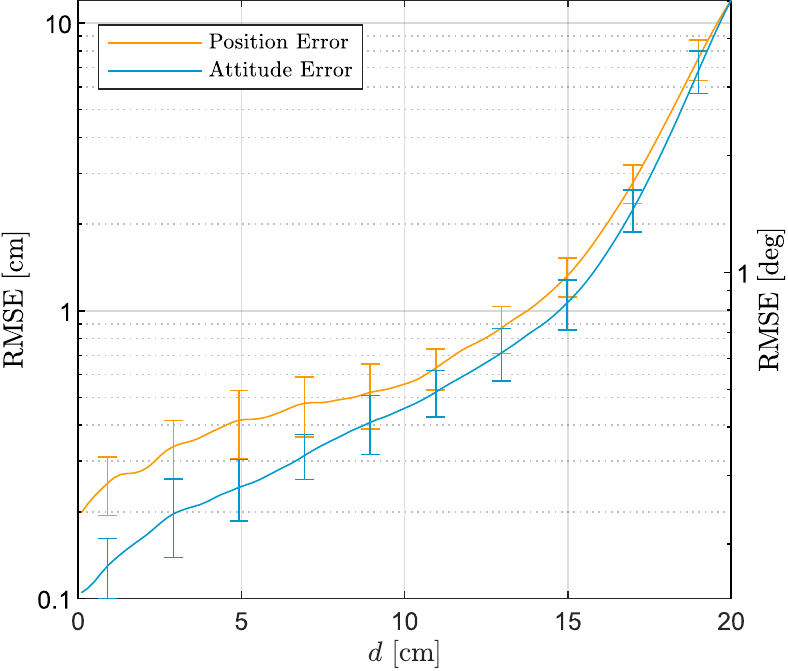}
	\caption{Pose estimation error in relation to distance $d$ from haptic sensor as RMSE (Root-Mean-Square-Error).}
	\label{cap_sensor_dist_err}
\end{figure}

\begin{figure}
	\centering
	\def\svgwidth{\columnwidth}
	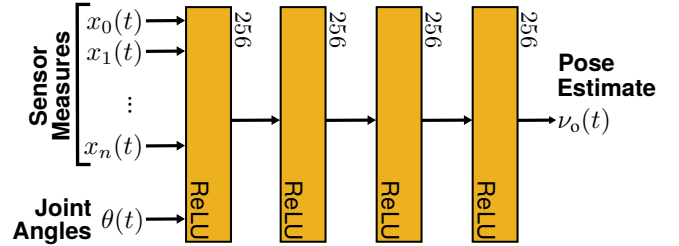
	\caption{Neural network using robot joint angles and sensor measures to generate object pose estimation. It contains four fully connected layers, each comprised of 256 ReLU units.}
	\label{fig:cap_sensor_neural_net}
\end{figure}

Using experimental data and the model derived in section \ref{sec:electric_model}, we generated a training dataset. In simulation, we mounted the sensors on the hull of a robot and recorded their response to nearby human limbs along with the robot's joint angles. 
In combination with the object pose this dataset, we can use the dataset to train a neural network on estimating pose according to sensor measures.
All relevant data can be found \href{https://github.com/Mike2208/prednet_localization_results}{here}.
The resulting pose estimation can be seen in Fig. \ref{cap_sensor_dist_err}. As shown, the sensor has an estimated sensing range of around 15cm. At greater distances, the pose estimation error increases exponentially.

	\subsection{Visual Modality} \label{sec:procnet}
	The visual modality uses a captured camera image and computes the pose of the considered object using its shape within the frame.
	
	\subsubsection{Perception Strategy}

\begin{figure}
	\centering
	\def\svgwidth{\columnwidth}
	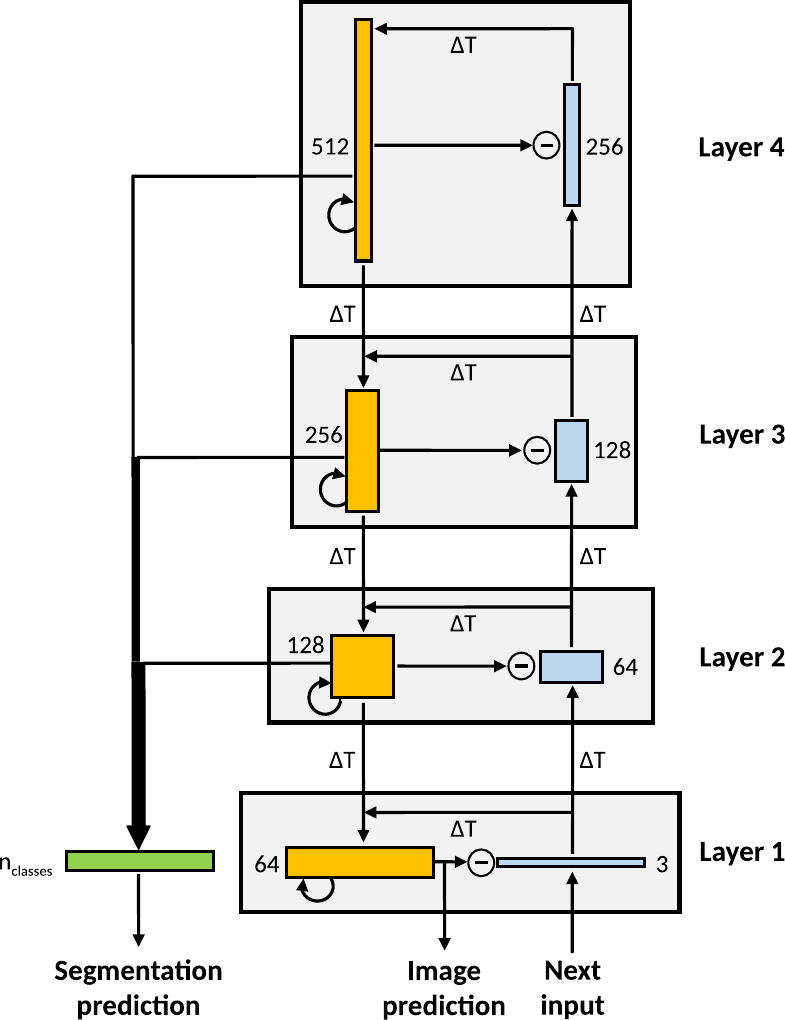
	\caption{Overview of the PredNet architecture. The width of each coloured box indicates the spatial scale of the layer component (the larger, the more resolution). Their height indicates how many feature maps are represented (also shown by each number). $\Delta t$ indicates connections for which axonal delays were implemented.}
	\label{fig:prednet_model_diag_a}
\end{figure}

\begin{figure}
	\centering
	\def\svgwidth{\columnwidth}
	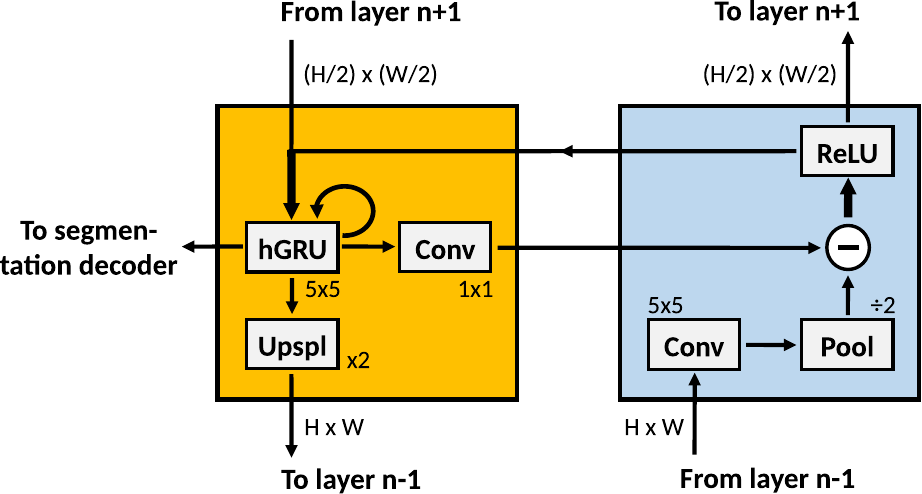
	\caption{Detailed view of the computations performed at each PredNet layer. The numbers represent kernel size for hGRU and Conv (convolution) operations, scaling factor for the Upspl (up-sample) and Pool operations. Note that Conv and Pool operations are absent in the first layer of PredNet. }
	\label{fig:prednet_model_diag_b}
\end{figure}

\begin{figure}
	\centering
	\def\svgwidth{\columnwidth}
	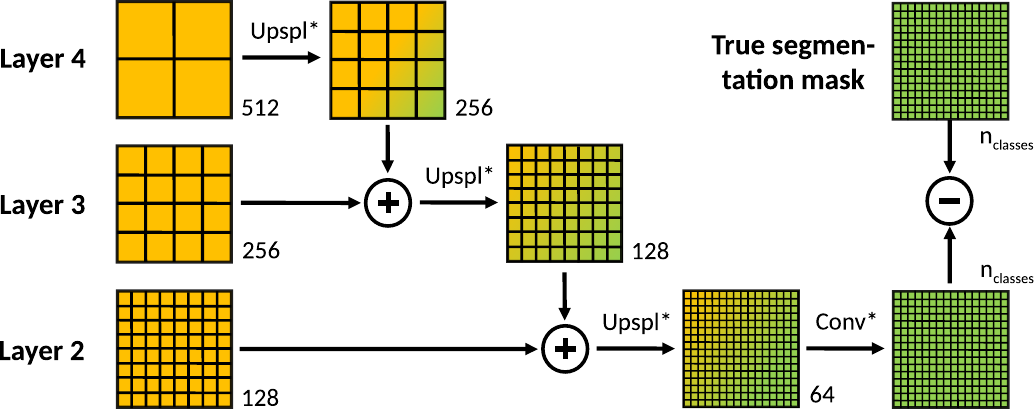
	\caption{Detailed view of how latent activity of the decoded layers representation components, coming with different resolutions, are combined in the segmentation prediction process. }
	\label{fig:prednet_model_diag_c}
\end{figure}

Detecting an object's pose in a camera frame requires two steps. First, we determine the portion of the frame that pertains to the target object. We use an approach derived from Predictive Coding methods to generate the corresponding segmentation mask that identifies an object's relevant data in the image. Then, we use a 3D representation of the object to determine the pose in fixed-frame coordinates. By translating and rotating the 3D representation to match the object's segmentation mask, we can estimate a pose corresponding to the object's true location. In the following, we will explain the procedure in detail.

\subsubsection{PredNet Model}
The initial segmentation mask is generated by a neural network based on PredNet architecture (see Fig. \ref{fig:prednet_model_diag_a}, \ref{fig:prednet_model_diag_b}, and \ref{fig:prednet_model_diag_c}). A PredNet is composed of a stack of layers that represent information and compute prediction error at their own level. Each layer is composed of a representation component (yellow boxes) and a prediction error computation component (blue boxes). The representation components use top-down information (arrows pointing downwards) as well as the error signal coming from their own layer (arrows pointing left) to generate an accurate prediction of the incoming bottom-up input of their own layer. They include recurrent units (circular arrows), in order to retain information for long periods of time. Going downstream, max pooling operations decrease spatial resolution, and convolutions increase the number of feature maps. The prediction error computation components take the difference between the prediction generated by the representation component at that layer (arrows pointing right), and the input received from the upstream layer (arrows pointing upwards). Their output constitutes the local prediction error signal of that layer. Image frames are sent to the first layer of the network and the prediction error signal is computed and propagated to the upstream layer, while top-down activity is projected downstream, based on the activity of the network related to the previous input frames. At every time step, the sum of all layers’ prediction error is computed and constitutes the self-supervised loss of the network. At the same time, the output of the representation component of all layers except the first one (which is used for image prediction), is sent to a segmentation decoding module (detailed computation in Figure 1C). The difference between the decoded segmentation mask and the true segmentation mask constitutes the supervised loss of the network. It is computed as the sum of the Dice loss [79] and the Focal loss [80]. An array of pixels representing the generated segmentation mask is
\begin{equation}
	m \in \mathds{N}^{h \times w},
\end{equation}%
with $h,w \in \mathds{N}$ as the height and width of the camera image in pixels. The values in $m$ correspond to the index of a detected object at the individual pixel coordinates, with a value of $0$ indicating no detected object.


\subsubsection{Vision-based Pose Estimation} 
Determining a fixed-frame pose from the segmentation mask is done with a 3D-representation of the object. As camera images are recorded at fixed time intervals $\Delta t_{\text c}$, assume that we are currently looking at frame $k \in \mathds{N}$ at time instance $t_k = k \Delta t_{\text c}$. We define
\begin{equation}
\begin{matrix}
	x_{\text{o}k} &= x_{\text{o}}(t_k), \\
	\Omega_{\text{o}k} &= \Omega_{\text{o}}(t_k),
\end{matrix}
\end{equation}%
with $x_{\text{o}k} \in \mathds{R}^3$ and $\Omega_{\text{o}k} \in \mathds{R}^3$ as the object pose and rotation at time $t_k$ in m and radians, respectively.
The current object pose can be determined by finding the translation and rotation for which the object shape most closely matches the segmentation mask $m_k$ at frame $k$. We use a 3D renderer based on the Godot-game engine to to determine 
\begin{equation}
	\hat{m}_k = f_{\text g}(\hat{x}_{\text{o}k}, \hat{\Omega}_{\text{o}k}),
\end{equation}%
where $\hat{x}_{\text{o}k}, \hat{\Omega}_{\text{o}k}$ represent an object's candidate pose, and $\hat{m}_k \in \mathds{N}^{h \times w}$ is the corresponding segmentation mask. The function $f_{\text g}(\centerdot)$ is used by the renderer to generate candidate segmentation masks from object poses. 
To determine the object's pose at frame $k$, we compare the rendered candidate mask with the segmentation mask determined by the PredNet as
\begin{gather}
\begin{aligned}
	\left[\begin{matrix}
		x_{\text{o}k} \\
		\Omega_{\text{o}k}
	\end{matrix}\right] &= \argmax_{\hat{x}_{\text{o}k}, \hat{\Omega}_{\text{o}k}} f_{\text m}(m_k, \hat{m}_k) \\
	&= \argmax_{\hat{x}_{\text{o}k}, \hat{\Omega}_{\text{o}k}} f_{\text m}(m_k, f_{\text g}(\hat{x}_{\text{o}k}, \hat{\Omega}_{\text{o}k})),
\end{aligned}
\end{gather}%
where $f_{\text m}(\centerdot)$ is a comparison function quantifying the overlap between two segmentation masks. For our use case, we assume the segmentation mask identifies the shape of a single object. Thus, the $m$'s entries are either 1 (if an object is present at the given pixel) or 0 (if no object is present). We set
\begin{equation}
	f_{\text m}(m, \hat{m}) = \sum_{i=1}^{h} \sum_{j=1}^{w} \frac{4(m_{ij} - \frac{1}{2}) (\hat{m}_{ij} - \frac{1}{2})}{wh},
\end{equation}%
with $m_{ij}, \hat{m}_{ij} \in [0,1]$ being the $i,j$ coefficients of $m$ and $\hat{m}$, respectively. This results in pixels where $m$ and $\hat{m}$ have equal values being weighted as 1, and pixels with unequal values weighted as -1. Therefore, the maximum value of $f_{\text m}(m, \hat{m})$ corresponds to the masks with the most overlap.
To decrease search times, we use information from the previous frame to compute the current object pose. In particular, starting from $x_{\text{o}k-1}$, $\Omega_{\text{o}k-1}$, a gradient ascent search on $f_{\text m}(m_k, \hat{m}_k)$ is performed to find the maximum overlap.
As $f_{\text m}(m_k, \hat{m}_k)$ represents the results of computing the shape of a non-trivial 3D-mesh in camera space, a direct derivation is not possible. Instead, we estimate the Jacbobian of $f_{\text m}(m_k, \hat{m}_k)$ in relation to $x_{\text{o}k}$ and $\Omega_{\text{o}k}$ by taking multiple samples around a point of interest. 
The Jacobian is then be used to follow the local gradient and determine the object's pose resulting in the most overlap between $m_k$ and $\hat{m}_k$. As we started the gradient ascent from the previous object's pose at $k-1$ and object motion between camera frames is fairly low, it is very unlikely that any local maxima (aside from the global one) are present in the search area. Images depicting the results of the consecutive steps are shown in Fig. \ref{fig:vis_segmentation}. The resulting modality offers accurate performance (less than 2cm error on average) up to a distance of around 5m when considering the entire object pose. Note however that at larger distances the main contributing factor to positional error is the relative distance to the camera; the modality can detect an object's relative pose up to a distance of around 10m when ignoring the relative distance (see the dashed orange line in Fig. \ref{fig:vis_segmentation}).
%

\begin{figure}
	\begin{center}
		\includegraphics[width=0.95\columnwidth]{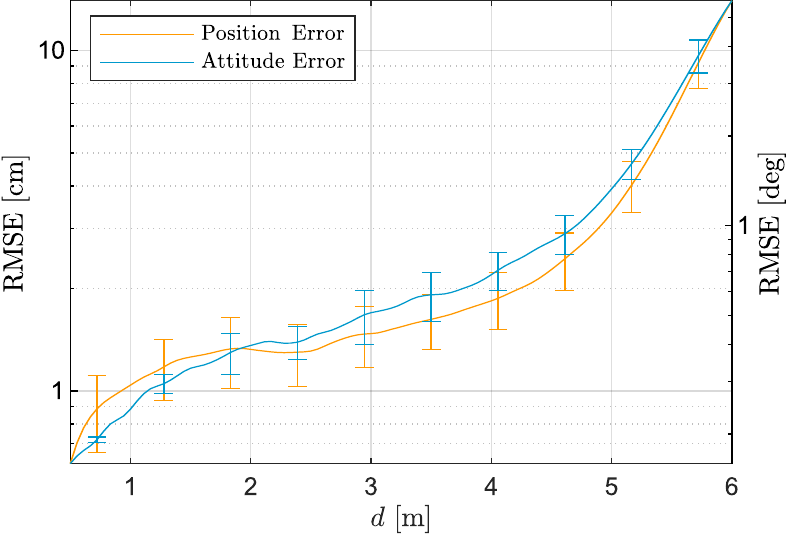}%
		\vspace{1em}
		\includegraphics[width=1\columnwidth]{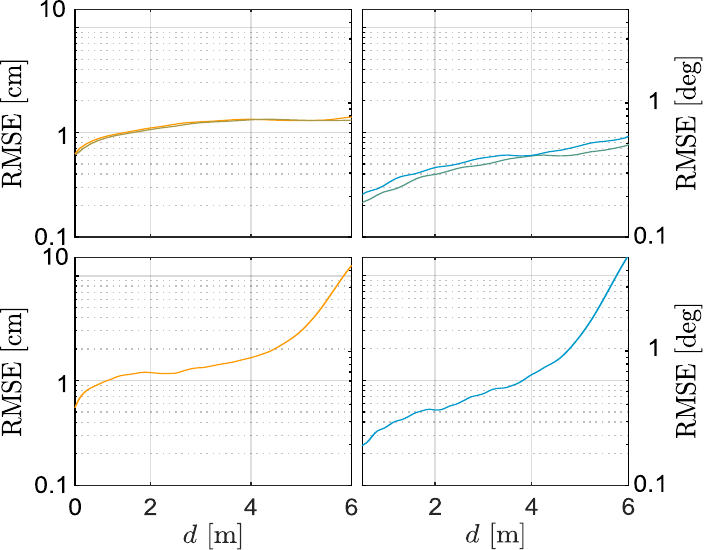}
	\end{center}
	\caption{Pose estimation error in relation to distance $d$ from camera as RMSE (Root-Mean-Square-Error). The subfigures show errors for the individual axes. The lower left figure depicts the xy-axis errors (x solid, y dashed) and RP-attitude.}
	\label{fig:vis_sensor_dist_err}
\end{figure}

\begin{figure}
	\centering
	\includegraphics[width=\columnwidth]{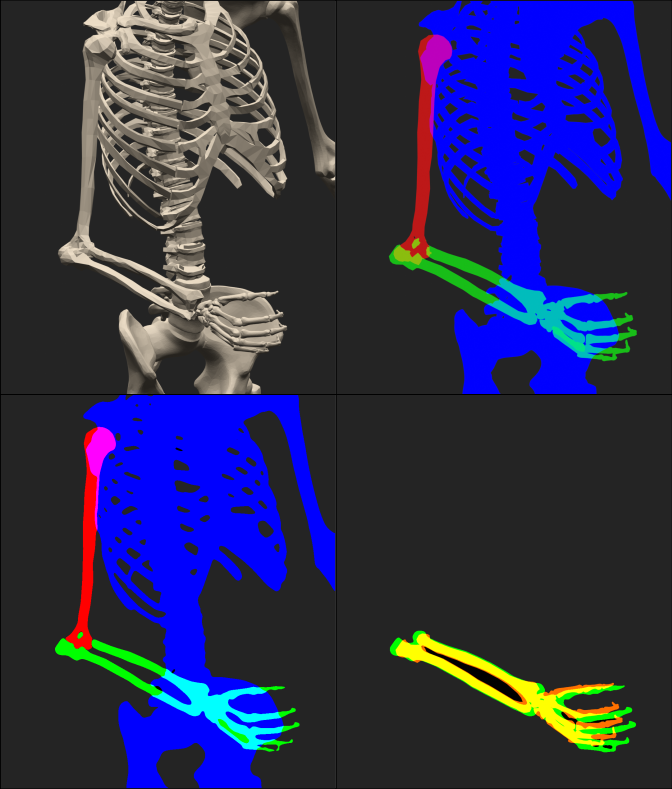}
	\caption{Segmentation and pose estimation of a human forearm. The top left image shows the camera frame. Top right and bottom left show the ground truth and the PredNet-generated segmentation masks, respectively. Each color denotes a separately segmented part of the body. Bottom right shows an example of pose estimation. The limb pose of the candidate mask that most closely matches the segmentation is selected.}
	\label{fig:vis_segmentation}
\end{figure}


	\section{Multimodal Localization} \label{sec:multimodal_localization}
	To combine the results of various sensors, multimodal localization requires the quantification of the quality of a sensor's performance. The pose returned by each modality can be weighed accordingly, with more accurate results being ranked higher than less precise ones.
	
	\subsection{Modified Luenberger-type Observer}
	We use a modified Luenberger observer model to combine the two previously introduced perception modalities.
To estimate the pose of a given object, we use a system defined as
\begin{align}
	\dot{x}(t)   &= f(x(t), u(t)) + n_{\text{x}}(t), \quad x(0) = x_0, \quad t \geqslant 0,\nonumber \\
	y_{\rm v}(t) &= g_{\rm v}(x(t)) + n_{\rm v}(t), \\
	y_{\rm h}(t) &= g_{\rm h}(x(t)) + n_{\rm h}(t), \nonumber
\end{align}%
with $x(t) \in \mathds{R}^6$ as the tracked object's pose in m and rad, $u(t) \in \mathds{R}^6$ as the input in m/s and rad/s, $n_{\text{x}}(t) \in \mathds{R}^6$ as the velocity noise in m/s and rad/s, $y_{\rm v}(t), y_{\rm h}(t) \in \mathds{R}^6$ as the pose in m and rad retrieved from the visual and haptic sensors, respectively, and $n_{\rm v}(t), n_{\rm h}(t) \in \mathds{R}^6$ as sensor noise in m and rad, respectively.
As the tracked object can move along arbitrary trajectories, $f(x(t), u(t))$ is a nonlinear function. Due to the nature of our sensors, $g_{\rm v}(x(t))$ and $g_{\rm h}(x(t))$ are likewise nonlinear.
For ease of exposition, we combine the two sensing modalities and define
\begin{align}
	y(t) &\triangleq g(x(t)) + n(t),\\
	g(x) &= \begin{bmatrix}
	g_{\rm v}(x)^{\rm T},
	g_{\rm h}(x)^{\rm T}
	\end{bmatrix}^{\rm T}, \nonumber\\
	n(t) &= \begin{bmatrix}
	n_{\rm v}(t)^{\rm T},
	n_{\rm h}(t)^{\rm T}
	\end{bmatrix}^{\rm T}. \nonumber
\end{align}%
To merge vision- and haptic-based pose estimations, we implement a modified Luenberger observer model as described in \cite{khalil2015nonlinear}, enabling us to track the object's trajectory. We use the observer variables
\begin{gather}
\begin{aligned}
	\dot{\hat{x}}(t) &= f(\hat{x}(t), u(t)) + G(t) z(t), &\hat{x}(0) &= \hat{x}_0,\\
	z(t) &= y(t) - \hat{y}(t),& t &\geqslant 0,\\
	\hat{y}(t) &= g(\hat{x}(t)) + n(t), &&
\end{aligned} \label{equ:luenberger_pose}
\end{gather}%
where $\dot{\hat{x}}(t) \in \mathds{R}^6$ is the estimated object pose, $\hat{y}(t) \in \mathds{R}^{12}$ is the pose measured by our perception modalities, $z(t) \in \mathds{R}^6$ is the pose error, and $G(t) \in \mathds{R}^{6\times 12}$ is the nonlinear observer gain. As the observer should track the object's true pose, our goal is to minimize the difference between actual pose $x(t)$ and estimated pose $\hat{x}(t)$, we define the error $e(t)$ as
\begin{gather}
\begin{aligned}
	e(t) &\triangleq x(t) - \hat{x}(t), \qquad t \geqslant 0,\\
	\dot{e}(t) &= f(x(t), u(t)) - f(x(t) - e(t)) \\
	&\quad- G(t) \left(y(t) - g(x(t) - e(t) + n(t)\right).
\end{aligned}
\end{gather}%
where $e(t) \in \mathds{R}^6$ is the observation error in m and rad. A Taylor series
expansion of $\dot{e}(t)$ at steady state $e(t_{\rm s}) = 0$ is performed, resulting in
\begin{align}
	\dot{e}(t) &= \Big(A(t) - G(t) C(t)\Big) e(t) + e_{\rm r}(t),
\end{align}%
with
\begin{align*}
	A(t) = \frac{\partial f(\hat{x}(t), u(t))}{\partial \hat{x}(t)}, \qquad C(t) = \frac{\partial g(\hat{x}(t)}{\partial \hat{x}(t)},
\end{align*}%
where $A(t) \in \mathds{R}^{6\times 6}$ and $C(t) \in \mathds{R}^{12\times 6}$ are the matrix linearizations of functions $f(\centerdot)$ and $g(\centerdot)$ at steady state $e(t_{\rm s}) = 0$, respectively.
	
\begin{align}
	K(t) =& P(t) C^{\rm T}(t) R^{-1}(t), \nonumber\\
	\dot{P}(t) =& A(t) P(t) + P(t) A^{\rm T}(t) + Q(t) \nonumber\\
	& - P(t) C^{\rm T}(t) R^{-1}(t) C(t) P(t), \label{equ:luenberger_estim}\\
	R(t) =& \text{diag}(R_{\text v}, R_{\text c}), \nonumber\\
	P(t_0) =& P_0, \nonumber
\end{align}%
where $Q(t) \in \mathds{R}^{6\times6}$, $R(t) \in \mathds{R}^{12\times12}$ are the noise covariance matrices of $n(t)$ and $m(t)$, respectively. $R_{\text v}, R_{\text c} \in \mathds{R}^{6 \times 6}$ are the noise covariance matrices of the vision- and haptic-based perception modalities, respectively.
$R_{\text v}$ and $R_{\text c}$ are estimated while individually for the two sensing modalities. In section \ref{sec:perception_methods_noise}, we detail how they are computed.
	
In our use cases, we found that the trajectories of tracked objects are smooth, with no external perturbations interrupting their movements. Therefore, we implemented a canonical 2nd order low-pass filter with $\omega_{\rm n} \in \mathds{R}^+$ as the angular cutoff frequency of the filter in rad/s, and $\zeta \in \mathds{R}^+$ is the filter ramp.
	
	\subsection{Integration of Perception Methods} \label{sec:perception_methods_noise}	
	\begin{figure*}[!t]
	\begin{minipage}[c]{\columnwidth}
		\includegraphics[width=\textwidth]{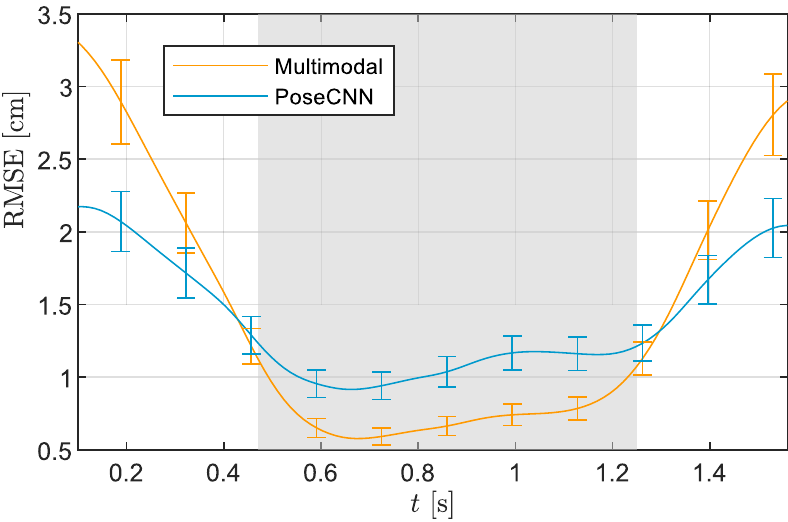}%
		\caption{Pose estimation error comparison between our multimodal approach (orange) and NVIDIA's PoseCNN (cyan).}
		\label{fig:loc_err_comp}
	\end{minipage}%
	\hfill%
	\begin{minipage}[c]{\columnwidth}
		\includegraphics[width=\textwidth, height=140pt]{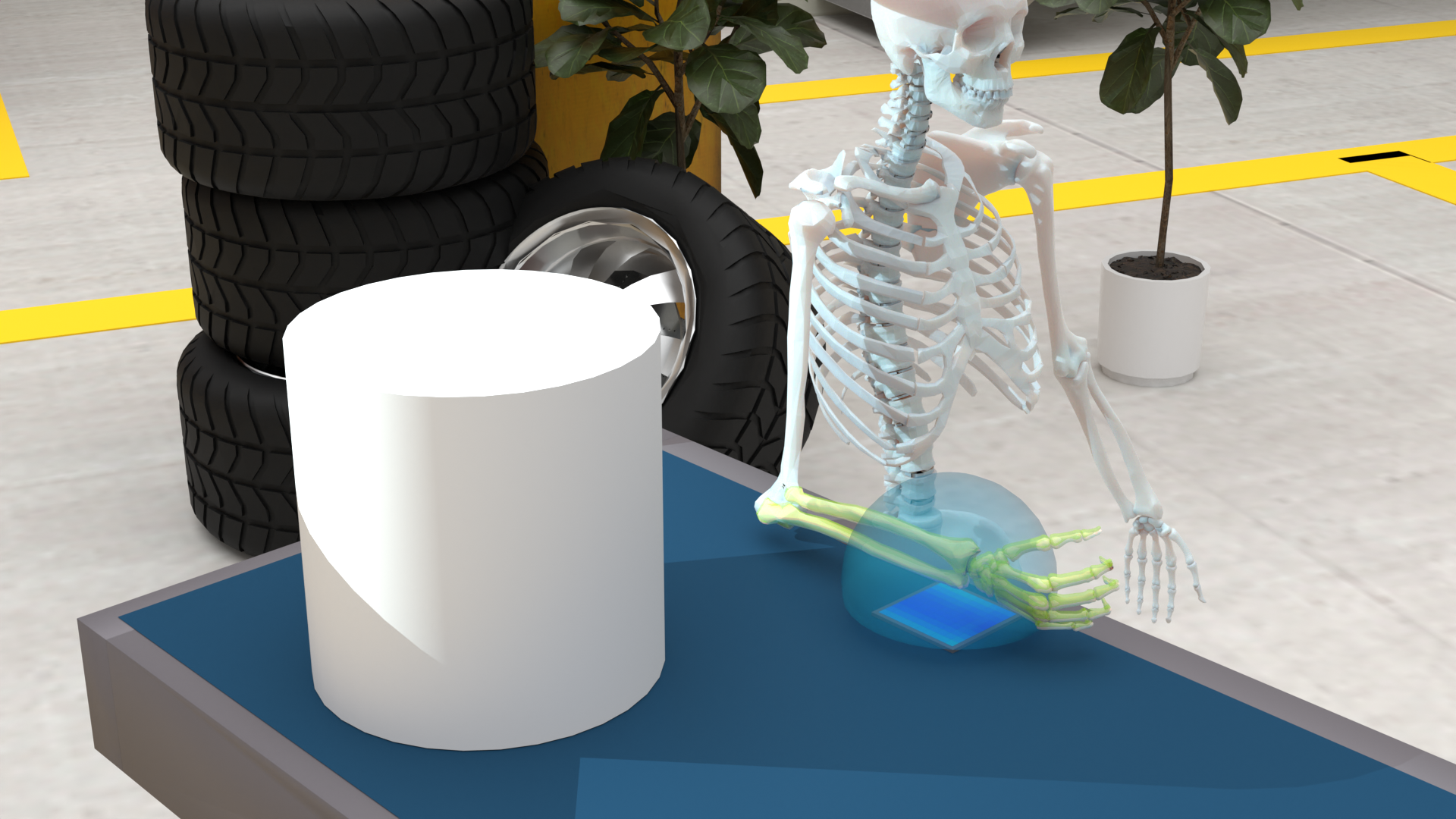}%
		\caption{Simulation environment for human limb tracking. The green forearm silhouette designates the estimated pose, the blue pads are the haptic sensors, and the illuminated areas represent the cameras field of view.}
		\label{fig:human_limb_environment}
	\end{minipage}%
\end{figure*}%

$R_{\text v}$ and $R_{\text c}$ are estimated individually for the two sensing modalities. For haptic sensors, we measured the noise covariance of a human forearm at various relative positions and orientations from the sensor, within sensing range of the electrodes (relative distance between object and electrode less than 15cm). We used Gaussian regression to determine a function representing the relation between measured relative pose and noise covariance, resulting in
\begin{equation}
	R_{\text c}(t) = f_{\text c}(\hat{x}(t), \theta(t)).
\end{equation}%
For vision-based localization, a different approach was implemented. We estimate $R_{\text v}(t)$ by determining the amount of occlusion; in general, the more an object is hidden, the less reliable vision-based pose estimation becomes, resulting in
\begin{equation}
R_{\text v}(t) = f_{\text v}(\hat{m}(t), m(t)),
\end{equation}%
with
\begin{equation}
	f_{\text v}(\hat{m}(t), m(t)) =  \frac{\sum_{i=1}^h \sum_{j=1}^w \hat{m}_{ij}(t) - m_{ij}(t)}{\sum_{i=1}^h \sum_{j=1}^w \hat{m}_{ij}(t)} W_{\text v},
\end{equation}%
where $W_{\text v} \in \mathds{R}^{6\times 6}$ is a diagonal matrix that determines the noise strength for each dimension. 

Having computed $R_{\text v}$ and $R_{\text c}$, we can use them to determine the influence that each sensing modality should have on the estimated pose. For instance, if an object is close to the haptic sensors, the sensor's SNR is high (due to the fact that the emitted electric field is heavily influenced by the close proximity of the scanned object). Therefore, the pose estimated by the haptic sensor should be more reliable than if an object were far away, and the sensor's measures should be weighted higher. This is reflected in how $R_{\text c}(t)$ influences the computation of $\hat{x}(t)$ in (\ref{equ:luenberger_estim}) and subsequently (\ref{equ:luenberger_pose}). Similarly, the pose determined by ProcNet is more reliable the less of an object is occluded, meaning its estimated pose should be weighted higher the less occlusion is observed. $f_{\text v}(\hat{m}(t), m(t))$ determines the level of occlusion by comparing the segmentation mask computed by our PredNet architecture with the one generated by the 3D-renderer. As the renderer ignores other objects in the scene, we can use it's segmentation mask to determine how much of an object is obstructed from the camera's view. We add an additional weight $W_{\text v}$ to designate how much occlusion influences the accuracy of vision-based pose estimation.

\begin{figure}[b]
	\includegraphics[width=\columnwidth]{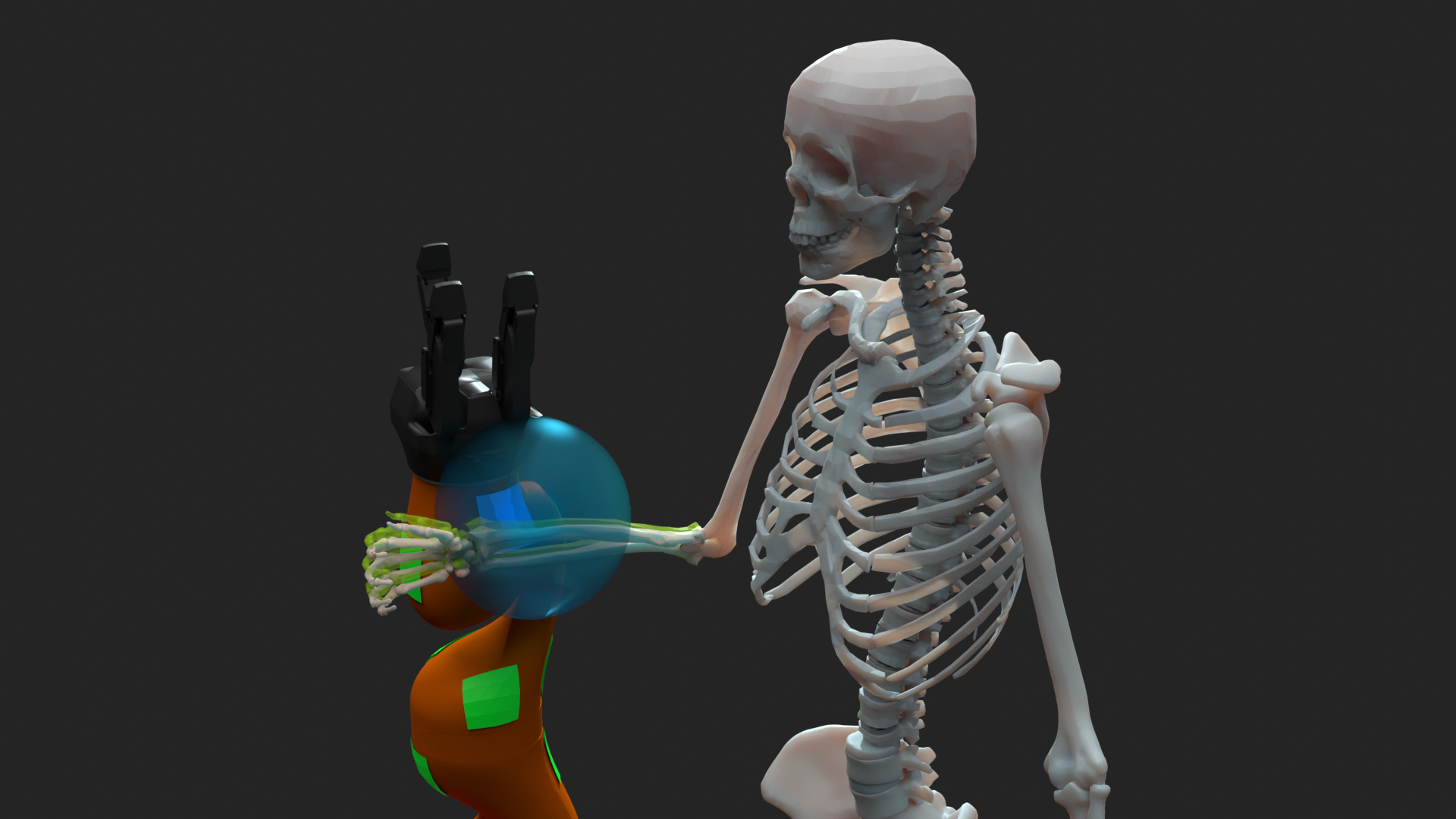}
	\caption{Human limb detection and pose estimation with mounted haptic sensors. The green silhouette designates estimated pose, and the green-blue pads show the haptic sensors.}
	\label{fig:robot_human_limb_environment}
\end{figure}
	
	\section{Numerical Simulation} \label{sec:human_limb_tracking}


	\begin{figure*}
	\centering
	\includegraphics[width=\textwidth]{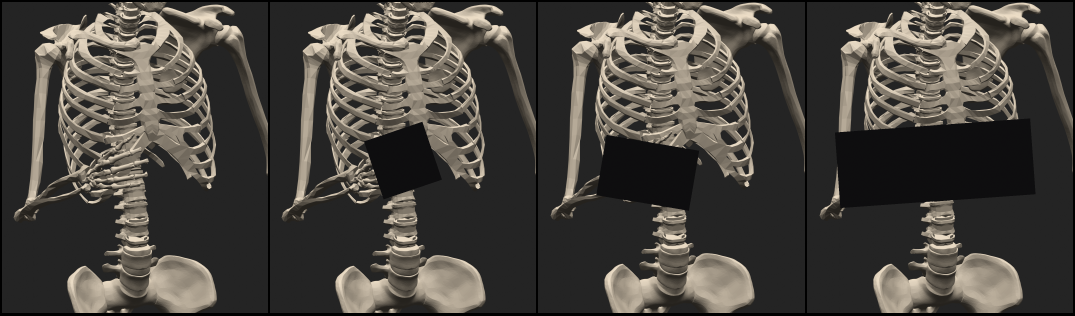}
	\caption{Example of various levels of occlusion (Top left: no occlusion, top right: light occlusion, bottom left: medium occlusion, bottom right: heavy occlusion).}
	\label{fig:occlusion_summary}
\end{figure*}

\begin{table*}
	\centering
	\begin{tabular}{|c|c|>{\centering\arraybackslash}p{6em}|>{\centering\arraybackslash}p{13em}||>{\centering\arraybackslash}p{6em}|>{\centering\arraybackslash}p{13em}|}
		\hline %
		\multicolumn{6}{|c|}{\textbf{Trajectories within range of haptic sensor}}\\
		\hline \hline %
		\textbf{Camera Distance} & \textbf{Occlusion Level} & ProcNet ($e_{\rm v}$) & Multimodal ProcNet ($e_{\rm m}$) & PoseCNN ($e_{\rm r}$) & Multimodal PoseCNN ($e_{\rm rm}$)\\
		\hline \hline %
		\multirow{3}*{Short (0--3m)} & Light (0--33\%)   & 1.73cm & \textbf{1.59cm} & 1.62cm & \textbf{1.59cm} \\
		& Medium (33--66\%)  & 2.48cm & \textbf{1.83cm} & 4.83cm & 2.59cm \\
		& Heavy  (66--100\%) & 5.84cm & \textbf{3.27cm} & 6.29cm & 3.37cm \\
		\hline %
		\multirow{3}*{Medium (3--6m)} & Light (0--33\%)  & 1.81cm & 1.32cm & 1.57cm & \textbf{1.28cm} \\
		& Medium (33--66\%)  & 2.59cm & \textbf{1.90cm} & 5.02cm & 2.18cm \\
		& Heavy (66--100\%) & 6.04cm & \textbf{3.54cm} & 6.54cm & 4.53cm \\
		\hline %
		\multirow{2}*{Long (6--10m)} & Light (0--33\%)  & 4.07cm & 3.06cm & 2.65cm & \textbf{2.43cm} \\
		& Medium (33--66\%) & 5.39cm & 4.32cm & 2.86cm & \textbf{2.90cm} \\
		\hline %
		\multicolumn{6}{l}{}\\
		\hline %
		\multicolumn{6}{|c|}{\textbf{Trajectories outside of range of haptic sensor}}\\
		\hline \hline %
		\textbf{Camera Distance} & \textbf{Occlusion Level} & ProcNet ($e_{\rm v}$) & Multimodal ProcNet ($e_{\rm m}$) & PoseCNN ($e_{\rm r}$) & Multimodal PoseCNN ($e_{\rm rm}$)\\
		\hline \hline
		\multirow{3}*{Short (0--3m)} & Light (0--33\%)  & 1.71cm & 1.71cm & 1.61cm & \textbf{1.61cm} \\
		& Medium (33--66\%) & 2.45cm & \textbf{2.45cm} & 4.79cm & \textbf{4.79cm} \\
		& Heavy (66--100\%) & 5.80cm & \textbf{5.80cm} & 6.25cm & 6.25cm \\
		\hline %
		\multirow{3}*{Medium (3--6m)} & Light (0--33\%) & 1.83cm & 1.83cm & 1.56cm & \textbf{1.56cm} \\
		& Medium (33--66\%) & 2.60cm & \textbf{2.60cm} & 5.03cm & 5.03cm \\
		& Heavy (66--100\%) & 5.99cm & \textbf{5.99cm} & 6.50cm & 6.50cm \\
		\hline %
		\multirow{2}*{Long (6--10m)} & Light (0--33\%)  & 3.98cm & 3.98cm & 2.61cm & \textbf{2.61cm} \\
		& Medium (33--66\%) & 5.33cm & 5.33cm & 2.85cm & \textbf{2.85cm} \\
		\hline
	\end{tabular}
	\caption{Average sensor error on relation to various scenarii. The pose errors of each individual sensing modality (PredNet-based vision as $e_{\rm v}$, and PoseCNN-based vision as $e_{\rm r}$) is listed. $e_{\rm m}$  and $e_{\rm rm}$ denote the pose errors of our multimodal method, using ProcNet and PoseCNN, respectively.}
	\label{tab:dataset_results}
\end{table*}%

A series of simulations were performed to determine the efficacy of our method, both with and without the presence of occlusion. First, we verified our setup in a straightforward scene; a human moves his forearm up and down over an array of haptic sensors (see Fig. \ref{fig:human_limb_environment}). An RGB camera is placed within the scene to deliver vision-based pose estimates via ProcNet. The setup is repeated multiple times, with varying levels of occlusion. The setup uses both sensing modalities to determine the pose of the human forearm. If the arm is too far away from the haptic sensors, pose estimation is performed by vision. Similarly, if view of the arm is obstructed (and it is in range of the sensor grid), pose estimation is performed by capacitive measures. In cases where both modalities are generating valid pose estimates, our observer model synthesizes the results. We compare our results to a pose estimation algorithm developed by NVIDIA, PoseCNN (\cite{xiang2017posecnn}).
A second set of simulations was performed with haptic sensors mounted on a robot arm (see Fig. \ref{fig:robot_human_limb_environment}). As previously described, a moving robot arm occludes nearby objects, leading to difficulties in pose estimation by vision-based methods alone. By mounting haptic sensors, we ensure that nearby objects can still be localized via a second method. Similarly to our previous scenario, we performed multiple tests with different levels of occlusion. The results can be seen in Fig. \ref{fig:loc_err_comp}. 
Model parameters for all simulations are configured as $W_{\text v} = \rm{diag}(8, 8, 8, 16\pi, 16\pi, 16\pi)$,  $\omega_{\rm n} = 120\pi$ and $\zeta = \frac{1}{\sqrt{2}}$.

\begin{figure*}
	\centering
	\def\svgwidth{\textwidth}
\begingroup%
  \makeatletter%
  \providecommand\color[2][]{%
    \errmessage{(Inkscape) Color is used for the text in Inkscape, but the package 'color.sty' is not loaded}%
    \renewcommand\color[2][]{}%
  }%
  \providecommand\transparent[1]{%
    \errmessage{(Inkscape) Transparency is used (non-zero) for the text in Inkscape, but the package 'transparent.sty' is not loaded}%
    \renewcommand\transparent[1]{}%
  }%
  \providecommand\rotatebox[2]{#2}%
  \newcommand*\fsize{\dimexpr\f@size pt\relax}%
  \newcommand*\lineheight[1]{\fontsize{\fsize}{#1\fsize}\selectfont}%
  \ifx\svgwidth\undefined%
    \setlength{\unitlength}{516bp}%
    \ifx\svgscale\undefined%
      \relax%
    \else%
      \setlength{\unitlength}{\unitlength * \real{\svgscale}}%
    \fi%
  \else%
    \setlength{\unitlength}{\svgwidth}%
  \fi%
  \global\let\svgwidth\undefined%
  \global\let\svgscale\undefined%
  \makeatother%
  \begin{picture}(1,0.33333333)%
    \lineheight{1}%
    \setlength\tabcolsep{0pt}%
    \put(0,0){\includegraphics[width=\unitlength,page=1]{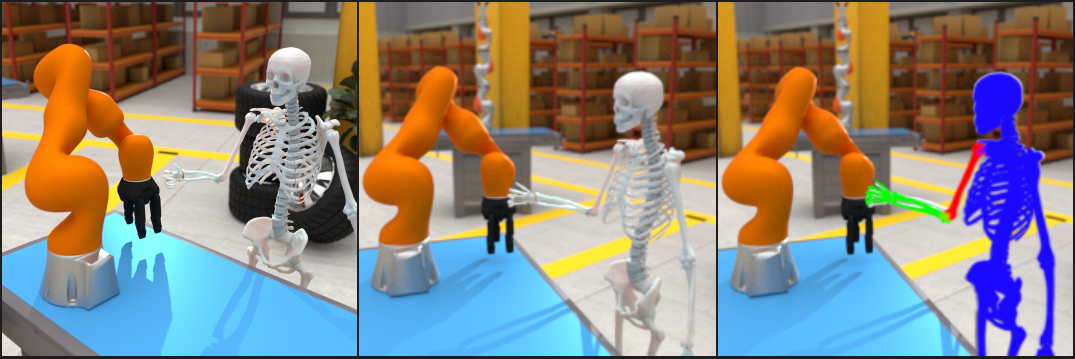}}%
    \put(0.66645378,0.28509751){\color[rgb]{1,1,1}\makebox(0,0)[t]{\lineheight{1.25}\smash{\begin{tabular}[t]{c}ProcNet\end{tabular}}}}%
    \put(0,0){\includegraphics[width=\unitlength,page=2]{cap_cam_results_03.pdf}}%
  \end{picture}%
\endgroup%

	\caption{Localization via visual modality. The left image shows an overview of the scene, the middle image is the camera input, and the right image depicts localized limb poses overlaid over the input image.}
	\label{fig:cap_cam_results}
\end{figure*}

Lastly, we designed a series of datasets to determine the system's performance under various degrees of occlusion and at varying distances from a camera (all relevant data can be found online \href{https://github.com/Mike2208/prednet_localization_results}{here}). The level of occlusion is determined by the amount of relevant pixels in the camera image that are obscured; here, the relevant pixels are those describing the shape of the considered objects. We therefore determine the degree of occlusion as the sum of all relevant yet obscured pixels divided by the sum of all pixels describing a considered object's shape. Thus, a frame with a completely visible object is noted as having 0\% occlusion, while a frame with a fully obscured object has 100\% occlusion. When evaluating various sensor distances, the level of occlusion was kept low, at less than 10\% (see Figure \ref{fig:occlusion_summary} for examples). Figure \ref{fig:cap_cam_results} shows an example of how individual modalities localize a human upper limb in our setup. Haptic sensors have a small detection range (around 20cm, see our previous work in \cite{zechmair2023active}) as denoted in the top right image. However, within this range, the sensor can provide sub-millimeter accuracy, as shown previously in Figure \ref{cap_sensor_dist_err}. ProcNet provides a larger sensing range, but also with higher errors. The bottom left image shows camera input, and the bottom right image shows the resultant localized human pose.
The results are shown in Table \ref{tab:dataset_results}. ProcNet outperforms PoseCNN when the camera is less than 6m away from the tracked object, starting at medium occlusion levels. When considering robot workcells, the setup is often constrained by similar parameters; a camera is placed within 6m of the robot, which is close enough to monitor the entire range of motion, but still far enough away to prevent collision. While moving, the robot will also generate a moderate amount of occlusion, thus preventing PoseCNN from performing accurate pose estimation. Another effect of occlusion is the marked performance increase of multimodal pose estimation, with this approach offering better results than the single-modality, vision-based pose estimation. Note that while haptic-based pose estimation is very accurate, the modality can only detect objects at close proximity.



	
	\section{Conclusion} \label{sec:conclusion}
%
	In this article, we described two sensing modalities, one based on vision, the other on electric perception. Using a Luenberger observer model, we combined the pose estimates generated by the individual sensors to deliver robust results in the presence of occlusion. In the future, we aim to improve usability of our setup. The model can be generalized to detect multiple targets and determine their respective poses instead of being limited to tracking single objects. Future work could also involve a more comprehensive method of estimating object poses based on haptic sensor measures. By taking the various surface points measured by individual sensors and interpreting them as a point cloud, a less heuristic approach to pose estimation could be considered.

	\hypersetup{urlcolor=black}
	\bibliographystyle{IEEEtran}%
	\bibliography{IEEEabrv,references}	
	
\end{document}